\documentclass[11pt]{article}

\usepackage{color}
\usepackage{soul}
\usepackage{cancel}

\usepackage{geometry}  % Flexible and complete interface to document dimensions
\usepackage[T1]{fontenc}  % Standard package for selecting font encodings
\usepackage[utf8x]{inputenc}  % Accept different input encodings
\usepackage{lineno}  % Line numbers on paragraphs
\usepackage{fancyhdr}  % Extensive control of page headers and footers in LaTeX2ε
\usepackage{enumitem}  % Control layout of itemize, enumerate, description
\usepackage{hyperref}  % Extensive support for hypertext in LaTeX
\usepackage{titling}  % Control over the typesetting of the \maketitle command
\usepackage{natbib}  % Flexible bibliography support
\usepackage{mathtools}  % Mathematical tools to use with amsmath
\usepackage{titlesec}  % Select alternative section titles
\usepackage{lastpage}  % Reference last page for Page N of M type footers
\usepackage{lscape}

\usepackage[english]{babel}  % Multilingual support for Plain TeX or LaTeX
\usepackage{amsmath}  % AMS mathematical facilities for LaTeX
\usepackage{amsfonts}  % TeX fonts from the American Mathematical Society
\usepackage{amssymb}  % Additional symbols from American Mathematical Society
\usepackage{wasysym}  % LaTeX support file to use the WASY2 fonts
\usepackage{bbm}  % "Blackboard-style" cm fonts
\usepackage{array}  % Extending the array and tabular environments
\usepackage{xr}  % References to other LaTeX documents
\usepackage{verbatim} % Reimplementation of and extensions to LaTeX verbatim
\usepackage{float} % Improved interface for floating objects
\usepackage{caption}
\usepackage{subcaption}
\usepackage{marginnote}
\usepackage[superscript,biblabel]{cite}

\usepackage{booktabs}


\hypersetup{%
  pdfborderstyle={/S/U/W 1}% border style will be underline of width 1pt
}

\setcitestyle{round,numbers}
\titleformat{name=\section}{\normalfont\Large\bfseries}{}{0pt}{}
\titleformat{name=\subsection}{\normalfont\large\bfseries}{}{0pt}{}
\titleformat{name=\subsubsection}{\normalfont\normalsize\bfseries}{}{0pt}{}

\newcommand{\email}[1]{\href{mailto:{#1}}{{#1}}}

\newcommand{\keywords}[1]{\textbf{Keywords}: {#1}}

\newcommand{\optincludegraphics}[2][]{}
\newcommand{\optinput}[1]{}
\newtagform{brackets}{[}{]}
\usetagform{brackets}

\newcommand{\thejournal}[1]{Magnetic Resonance in Medicine}

\title{Contrast-Informed Augmentation and Domain-Adversarial Training for Adult-to-Neonatal MR Reconstruction Generalization}
\begin{document}

% ======================================================================
%TC:ignore
\begin{titlepage}
{\noindent\LARGE\bf \thetitle}

\bigskip

\begin{flushleft}\large
    Stephen Moore\textsuperscript{1-3},
    Lara Leijser\textsuperscript{3-5},
    Richard Frayne\textsuperscript{1-3,6},
	Roberto Souza\textsuperscript{3,7,*}
\end{flushleft}

\bigskip

\noindent

\begin{enumerate}[label=\textbf{\arabic*}]
\item Biomedical Engineering, University of Calgary, Calgary, Alberta
\item Seaman Family MR Research Centre, Foothills Medical Centre, Calgary, Alberta
\item Hotchkiss Brain Institute, University of Calgary, Calgary, Alberta
\item Pediatrics, Division of Neonatology, University of Calgary, Calgary, Alberta
\item Alberta Children's Hospital Research Institute, University of Calgary, Calgary, Alberta
\item Radiology and Clinical Neuroscience, University of Calgary, Calgary, Alberta
\item Electrical and Software Engineering, University of Calgary, Calgary, Alberta

\end{enumerate}

\bigskip

\textbf{*} Corresponding author:

\indent\indent
\begin{tabular}{>{\bfseries}rl}
Name		&  Roberto Souza, PhD													\\
Department	& Electrical and Software Engineering													\\
Institution	& University of Calgary														\\
Address 	 & ICT 352C \\
            & 2500 University Dr NW													\\
			& Calgary, Alberta														\\
            & Canada	T2N 1N4  													\\
E-mail		& \email{roberto.souza2@ucalgary.ca}											\\
\end{tabular}

\vfill
%\wordcount{4147}{248}
% ======================================================================

\end{titlepage}
%TC:endignore
% ======================================================================

% ======================================================================
% ======================================================================
\pagebreak
% ======================================================================
% ======================================================================

% ======================================================================
%TC:break Abstract
\begin{abstract}

\textbf{Purpose}: 

To investigate whether contrast-informed data augmentation and domain-adversarial training improve the adult-to-neonatal generalization of the End-to-End Variational Network.

\textbf{Methods}: 

Three training regimes were investigated: (1) adult-only training with unaugmented adult data, (2) mixed training with paired unaugmented and neonatal-informed augmented adult data, and (3) mixed training with a domain-adversarial objective. Models were trained on retrospectively undersampled multi-coil adult T2-weighted brain MR data and evaluated on neonatal and adult test data at acceleration factors $R=4$ and $R=8$ using quantitative metrics and qualitative evaluation. Feature analyses assessed whether domain-adversarial training altered the latent representations of unaugmented adult, augmented adult, and neonatal test samples.

\textbf{Results}: 

Mixed training (Mixed) and mixed domain-adversarial training (Mixed-DAT) outperformed unaugmented adult-only training (Unaug-Only) when evaluated on neonatal data. At $R=4$, Mixed-DAT achieved the best performance ($\text{SSIM} = 0.924 \pm 0.027$, $\text{PSNR} = 33.98 \pm 1.15 \text{ dB}$). At $R=8$, Mixed-DAT performed best when measured using SSIM ($0.848 \pm 0.031$ vs. $0.766 \pm 0.037$ for Unaug-Only and $0.814 \pm 0.035$ for Mixed) and Mixed performed best when measured using PSNR ($29.56 \pm 0.83 \text{ dB}$ vs. $26.26\pm0.78 \text{ dB}$ for Unaug-Only and $29.43 \pm 0.83 \text{ dB}$ for Mixed-DAT). Qualitative assessment of t-SNE plots suggested that Mixed-DAT increased the overlap among the latent representations of the unaugmented adult, augmented adult, and neonatal test data.

\textbf{Conclusion}: 

Contrast-informed augmentation and domain-adversarial training improved adult-to-neonatal generalization of deep learning-based MR reconstruction. These findings suggest that contrast-informed data augmentation combined with adversarial training may improve robustness to domain shift in undersampled neonatal MR reconstruction.

\end{abstract}

%% RF: I am going to stop edtting MRI to MR - but i think this needs some thought.
%% SM: Swapped.

% ======================================================================
% : set search-engine keywords (3 to 6)
\bigskip

%TC:break _main_
% ======================================================================
% ======================================================================
\pagebreak
% ======================================================================
% ======================================================================

% ======================================================================
\section{Introduction}
% ======================================================================

Magnetic resonance (MR) imaging produces high-quality, high-contrast images for both clinical and research applications \cite{larkman_parallel_2007}. However, its long acquisition time--often exceeding 30 minutes in clinical settings--increases patient discomfort and the likelihood of motion-related artifacts, particularly in neonates \cite{larkman_parallel_2007,nguyen_prevalence_2020}. It also decreases patient throughput and, consequently, accessibility \cite{nguyen_prevalence_2020}. To mitigate this limitation, acceleration methods, such as parallel imaging (PI), undersample \textbf{k}-space data during acquisition \cite{larkman_parallel_2007}. This undersampling makes the recovery of high-quality images a more challenging inverse problem, thereby complicating reconstruction \cite{larkman_parallel_2007}. Conventional physics-informed algorithms, including SMASH \cite{sodickson_simultaneous_1997}, SENSE \cite{pruessmann_sense_1999}, and GRAPPA \cite{griswold_generalized_2002}, reconstruct undersampled PI data but are limited to relatively modest acceleration factors $R$, beyond which their performance tends to deteriorate. Here, $R$ denotes the degree of \textbf{k}-space undersampling and the corresponding reduction in acquisition time \cite{larkman_parallel_2007}.

Deep learning (DL)-based MR reconstruction methods have emerged as alternatives to conventional approaches \cite{knoll_deep-learning_2020}, demonstrating improved reconstruction of highly undersampled data (\textit{i.e.}, $R \ge 4$) \cite{muckley2021results} and robustness to overlapping receiver coil geometries \cite{dubljevic2024effect}. However, concerns about generalizability, referring to their ability to perform well on data outside their training distribution, impede their widespread clinical adoption \cite{knoll_deep-learning_2020}. Because these models are data-driven, they learn statistical priors over the training distribution. They may therefore fail under domain shift induced by differences in contrast, anatomy, or signal-to-noise ratio (SNR) \cite{knoll_deep-learning_2020}. Moreover, publicly available MR reconstruction datasets--specifically, datasets that contain raw multi-coil \textbf{k}-space data--are relatively scarce and consist primarily of healthy adult data, further restricting generalization to underrepresented domains.

Neonatal imaging is one such underrepresented domain. To our knowledge, no publicly available MR reconstruction datasets contain neonatal data. Large image-domain neonatal MR imaging datasets exist (\textit{e.g.}, the Developing Human Connectome Project (dHCP) \cite{edwards_developing_2022}), but are not suitable to evaluate DL-based MR reconstruction for the reasons outlined in Shimron et al.\cite{shimron_implicit_2022}. The generalizability of these reconstruction models to neonatal data is of particular relevance because characteristics of neonatal MR, including contrast inversion between white and grey matter, differences in anatomical scale, and reduced SNR, have been reported to impair generalizability \cite{knoll_assessment_2019,huang_evaluation_2022}. These differences are particularly relevant to DL-based MR reconstruction because learned reconstruction networks rely on anatomical and contrast-dependent image priors, which may not reliably transfer from adult to neonatal anatomy. Given the importance of neonatal MR imaging in developmental research, injury diagnosis, neurodevelopmental outcome prediction, and longitudinal monitoring, ensuring robust model performance in this domain is critical before any clinical adoption \cite{batalle_annual_2018,dubois_mri_2021}. However, the scarcity of neonatal data motivates approaches that better leverage comparatively abundant adult data.

This study investigated whether domain-adversarial training improves the generalizability of deep learning-based MR reconstruction models from adult to neonatal data. Domain adversarial training has been widely studied in domain generalization and applied to related medical imaging generalization problems such as segmentation and prediction \cite{ganin_domain-adversarial_2016,kamnitsas_unsupervised_2017,omidi2025improving}. However, its application in MR reconstruction, and particularly in adult-to-neonatal generalization, remains limited. We trained the commonly used End-to-End Variational Network (E2E-VarNet) \cite{sriram2020end}, which has performed well in MR reconstruction challenges \cite{muckley2021results,beauferris2022multi}, using pairs of unaugmented and augmented adult data, with and without an adversarial domain-classification objective, and evaluated reconstruction performance on held-out neonatal and adult test data. We augmented the adult data using a contrast-informed framework designed to approximate neonatal MR characteristics. We hypothesized that combining contrast-informed augmentation with domain-adversarial training would improve neonatal reconstruction performance while preserving adult-domain performance. Finally, we examined whether domain-adversarial training altered the learned latent representation to reduce domain separability.

The principal contributions of this study are: (1) evaluating contrast-informed adult data augmentation for improving generalization to neonatal MR reconstruction data; (2) incorporating domain-adversarial training into a DL-based MR reconstruction model to reduce sensitivity to augmentation-defined domain differences; and (3) assessing whether reconstruction improvements correspond to reduced latent feature separability between unaugmented adult, augmented adult, and neonatal test samples.

% ======================================================================
% \section{Theory}
% ======================================================================

% ======================================================================
\section{Methods}
% ======================================================================

\subsection{Overview of Experimental Design}

We investigated the effect of data augmentation and domain-adversarial training (DAT) on the generalizability of DL-based MR reconstruction models to out-of-domain (OOD) neonatal data. We considered three training regimes: (1) training on unaugmented adult data only, (2) training on a mixture of unaugmented and augmented adult data, and (3) training on the same mixed dataset with an additional DAT objective. 

%% RF: Reorder the presentation in this paragraphy below to refelect how the regimes are fist introduced YOu now go 3-2-1!
%% SM: Fixed.

The adult-only regime, which produced the model hereafter termed Unaug-Only, was trained with exclusively unaugmented adult data as a control. The mixed regime, which produced the model hereafter termed Mixed, was trained using paired unaugmented and augmented samples generated from the same adult training slices. Finally, the DAT regime, which produced the model hereafter termed Mixed-DAT, was trained on the same mixed dataset, but with an additional domain classification objective. Model training thus minimized a reconstruction loss and maximized a domain classification loss.  

All models were evaluated on a held-out neonatal test set. No neonatal data were used for model training, validation, or hyperparameter selection. We hypothesized that models trained with DAT would learn more domain-invariant representations and exhibit improved generalization to neonatal data compared to both controls. Thus, we evaluated each model for their reconstruction generalization to neonatal data and domain separability of latent representations.

\subsection{Datasets}

\subsubsection{Adult Dataset}

The adult dataset, used exclusively for model training, consisted of raw, fully sampled multi-coil \textbf{k}-space data from $n=55$ T2-weighted brain MR volumes obtained from the fastMRI dataset \cite{knoll_fastmri_2020}. We used the multi-coil brain subset and restricted the data to T2-weighted acquisitions to better align with the neonatal data. The training dataset had 50 volumes, and the validation dataset had 5. For each volume, we retained the complex-valued multi-coil \textbf{k}-space data, estimated coil sensitivity profiles using ESPIRiT\cite{uecker2014espirit}, and generated ground-truth images using root-sum-of-squares (RSS) reconstruction from the fully sampled data. These ESPIRiT-estimated coil sensitivity profiles were used exclusively for data augmentation. The reconstruction network did not receive externally estimated sensitivity profiles during training or inference.

%% RF: Define the parameters in alpha? N_sl, f?
We preprocessed the data by volume; however, because E2E-VarNet is a 2D network, we trained and evaluated the models by slice. During training and evaluation, we normalized the input \textbf{k}-space data as in the original E2E-VarNet paper \cite{sriram2020end}. Specifically, the data were scaled by factor $\alpha = \frac{\sqrt{n_{\text{sl}}} \cdot 10^4}{\|f\|_2}$, where $n_\text{sl}$ is the number of slices in the volume and $\|f\|_2$ is the $L^2$ norm. Undersampling masks corresponding to acceleration factors of $R=4$ or $R=8$, depending on the experiment, were applied retrospectively during training and inference. 2D equidistant Cartesian undersampling masks, applied slice-wise, were used to reflect the 2D acquisition of the data. By convention, the central 28 lines in \textbf{k}-space were left fully sampled for the auto-calibration signal (ACS). 

\subsubsection{Neonatal Dataset}

The neonatal dataset, used exclusively for model evaluation, consisted of raw, fully-sampled multi-coil \textbf{k}-space data from $n=30$ T2-weighted brain MR volumes (axial T2 FSE, 3 mm slice thickness) acquired at the Alberta Children's Hospital as part of the P3 Cohort study \cite{pekarsky_social_2022}. Ground-truth image generation and data preprocessing of the neonatal data mirrored those of the adult data, save for the estimation of coil sensitivity profiles, as the neonatal data were not augmented, and the removal of zero-padding from the raw \textbf{k}-space data, as zero-padding can artificially inflate reconstruction metrics \cite{shimron_implicit_2022}.  

\subsection{Data Augmentation and Domain Shift Simulation}

We designed and implemented a contrast-informed augmentation protocol to liken adult data to neonatal data. We applied this augmentation protocol at the slice level before data normalization. Adult multi-channel data were transformed to image space and coil-combined using sensitivity profiles estimated by ESPIRiT \cite{uecker2014espirit}. The pixels with the lowest intensity, primarily corresponding to the background or skull tissue, were excluded before \textbf{k}-means clustering ($k=3$) partitioned images by intensity to approximate white matter, grey matter, and cerebrospinal fluid (CSF). Masks approximating white matter, grey matter, and CSF were generated from these intensity partitions. This partitioning included the subcutaneous and subgaleal layers of the scalp, but these tissues were typically clustered with the CSF, which was not involved in the subsequent intensity-altering augmentations. We inverted the white and grey matter intensities to mimic neonatal T2-weighted image contrast and applied spatial smoothing to soften tissue boundaries. White and grey matter intensities were then rescaled to better approximate neonatal image characteristics, and the images were projected back into multiple channels using the estimated sensitivity profiles and transformed back to \textbf{k}-space for model input. 

\subsection{Reconstruction Model Training}

These experiments used the E2E-VarNet as the reconstruction network \cite{sriram2020end}. Architecturally, it consists of a cascade of 12 iterative blocks, each comprising a data consistency module and a learned regularization module implemented as a convolutional neural network. It also learns sensitivity profiles implicitly during model training. A Structural Similarity Index Measure \cite{wang_image_2004} (SSIM)-based loss function ($1-\text{SSIM}$) was minimized by an Adam optimizer with an initial learning rate of $1 \times 10^{-3}$ that decreased to $1 \times 10^{-4}$ after 40 epochs \cite{kingma_adam_2017}. The models were trained with a batch size of 1, necessitated by the varying dimensions of the training data, but using gradient accumulation to simulate a batch size of 16.

\subsection{Domain-Adversarial Training}

\begin{figure}
    \centering
    \includegraphics[width=\linewidth]{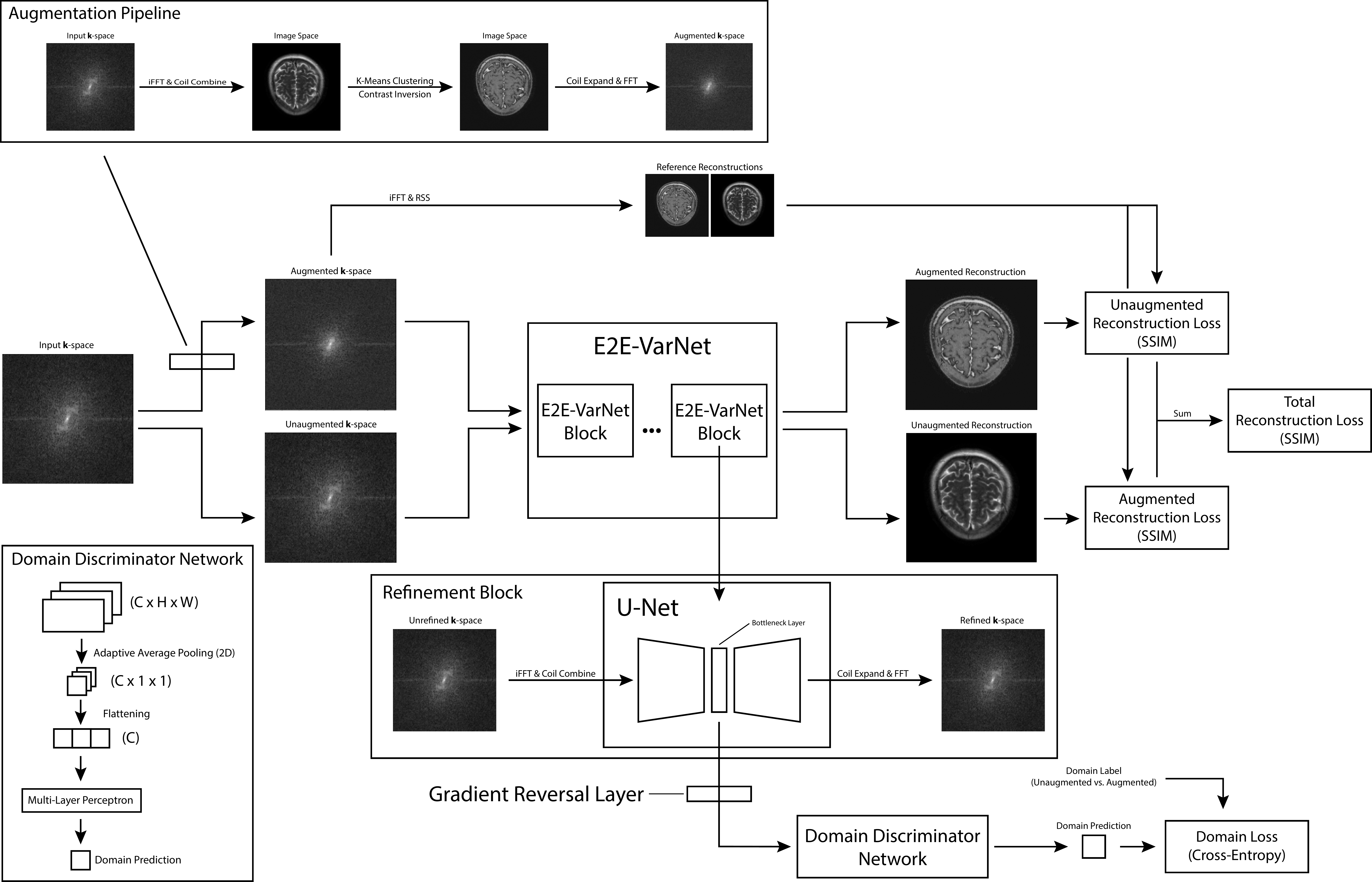}   
    \linespread{1}
    \caption{
        Domain-adversarial training pipeline. Paired unaugmented and neonatal-like augmented adult \textbf{k}-space samples were generated from the same training slice and passed through the E2E-VarNet reconstruction model \cite{sriram2020end}. In the Mixed-DAT regime, latent features from the bottleneck of the final refinement module were passed through a gradient reversal layer to a domain discriminator trained to classify samples as unaugmented or augmented. FFT and iFFT denote fast Fourier transform and inverse fast Fourier transform, respectively. RSS denotes root sum of squares. SSIM denotes structural similarity index measure \cite{wang_image_2004}.
    }
    \label{fig:methods-diagram}
\end{figure}

We implemented DAT by attaching a domain discriminator to the E2E-VarNet at the level of its latent feature representation (Figure \ref{fig:methods-diagram}). Specifically, we extracted these features from the bottleneck layer of the final refinement module, applied a 2D adaptive average pooling operation per channel, and flattened the result into a 1D tensor before passing it to the discriminator module. The discriminator module consisted of an input layer, a hidden layer, and an output layer, each separated by leaky ReLU activation functions. We inserted a gradient reversal layer (GRL) between the discriminator module and the rest of the reconstruction network to enable adversarial training. The GRL acts as an identity function during forward propagation but multiplies gradients by the negative of a scalar, denoted $\lambda$, during backpropagation. This multiplication pushes feature extractors--in our implementation, the E2E-VarNet up to the final refinement module's bottleneck layer--to learn representations that maximize a domain classifier loss, thereby enabling domain-invariant feature learning. $\lambda$ is an adversarial weighting parameter, controlling the weighting of the domain classification loss with respect to the reconstruction loss during the optimization of the shared parameters in the reconstruction network.

Thus, during training, optimization of the reconstruction network both minimized the reconstruction loss and maximized the domain classification loss, in principle learning a model that produces high-quality reconstructions from domain-invariant feature representations. We used cross-entropy for the domain classification loss and SSIM for the reconstruction loss. We used the same $\lambda$ scheduler as described in Ganin et al.\cite{ganin_domain-adversarial_2016}: $\lambda_p=\frac{2}{1+e^{-10p}}-1$, where $p$ denotes normalized training progress (\textit{i.e.}, a given training epoch as a fraction of the maximum number of training epochs). Domain labels were assigned according to whether inputs were unaugmented or augmented. Training batches comprised the unaugmented and augmented data pair generated from the same image slice. Otherwise, DAT used the same optimizer, learning rate, and learning rate schedule as the non-DAT regime.

\subsection{Training and Evaluation Protocols}

Three models--Unaug-Only, Mixed, and Mixed-DAT--were produced from the regimes. Unaug-Only was trained exclusively on unaugmented adult data, Mixed on a combination of unaugmented and augmented adult data, and Mixed-DAT on the same combination of unaugmented and augmented adult data with the addition of the DAT objective. For the regimes that produced Mixed and Mixed-DAT, we generated unaugmented and augmented data pairs from each slice and included these together in each training batch. All models shared identical reconstruction 
architectures, preprocessing, and optimization settings to isolate the effect of data composition and the use of DAT on neonatal generalization.

We evaluated all models using the neonatal test set to measure OOD generalizability. Reconstruction performance was quantified using SSIM and peak signal-to-noise ratio (PSNR). The metrics were computed by slice and averaged at the volume level for statistical comparison. To assess statistical significance, we performed a Friedman test followed by \textit{post hoc} Wilcoxon signed-rank tests with a Bonferroni correction. LL, a clinician-scientist specialized in neonatology, performed a qualitative evaluation of image quality.

To evaluate the effect of DAT on learned feature representations, we extracted latent features from the bottleneck layer of the final refinement module for unaugmented adult, augmented adult, and neonatal test samples. Features were extracted using the same layer provided to the domain discriminator, globally pooled using adaptive average pooling, and flattened into one-dimensional feature vectors. We then visualized the resulting feature distributions using t-distributed stochastic neighbor embedding (t-SNE).\cite{van2008visualizing} This analysis was used to qualitatively assess whether DAT reduced separability between unaugmented and augmented adult samples, and whether neonatal test samples occupied a more overlapping region of the learned feature space. Because t-SNE is a nonlinear visualization method, these analyses were interpreted qualitatively and not used as direct evidence of a causal mechanism.

\subsection{Implementation Details}

All experiments were conducted on a single NVIDIA L40 GPU. Training and evaluation were implemented in PyTorch using custom code made available at \url{https://github.com/moorestephen/unaug-aug-dat-mr-recon}. Random seeds were fixed for reproducibility.

% ======================================================================
\section{Results}
% ======================================================================

\subsection{Quantitative Reconstruction Performance}

\begin{table}[ht]
    \resizebox{\textwidth}{!}{
        \begin{tabular}{lcccccccc}
        \toprule
        & \multicolumn{4}{c}{Neonatal} & \multicolumn{4}{c}{Adult} \\
        \cmidrule(lr){2-5} \cmidrule(lr){6-9}
        Model 
        & R4 SSIM & R4 PSNR & R8 SSIM & R8 PSNR 
        & R4 SSIM & R4 PSNR & R8 SSIM & R8 PSNR \\
        \midrule
        Unaug-Only 
        & 0.866 $\pm$ 0.032 & 32.68 $\pm$ 1.18 & 0.766 $\pm$ 0.037 & 26.26 $\pm$ 0.78
        & 0.938 $\pm$ 0.034 & 38.03 $\pm$ 2.43 & \textbf{0.917 $\pm$ 0.035} & 34.11 $\pm$ 1.55 \\
        Mixed
        & 0.895 $\pm$ 0.027 & 33.49 $\pm$ 1.14 & 0.814 $\pm$ 0.035 & \textbf{29.56 $\pm$ 0.83}
        & \textbf{0.940 $\pm$ 0.032} & \textbf{38.37 $\pm$ 2.32} & 0.916 $\pm$ 0.038 & \textbf{34.28 $\pm$ 2.00} \\
        Mixed-DAT
        & \textbf{0.924 $\pm$ 0.027} & \textbf{33.98 $\pm$ 1.15} & \textbf{0.848 $\pm$ 0.031} & 29.43 $\pm$ 0.83
        & 0.937 $\pm$ 0.032 & 35.29 $\pm$ 2.06 & 0.908 $\pm$ 0.037 & 33.05 $\pm$ 1.54 \\
        \bottomrule
        \end{tabular}
    }
    \caption{Reconstruction performance, reported as mean $\pm$ standard deviation across test volumes. Bold values indicate the highest mean performance for each domain, acceleration factor, and metric. Statistical comparisons %were performed using Friedman tests followed by Bonferroni-corrected Wilcoxon signed-rank tests and 
    are summarized in Figure \ref{fig:reconstruction-metrics-boxplots}.}
    \label{tab:results_combined}
\end{table}

\begin{figure}
    \centering
    \includegraphics[width=0.99\linewidth]{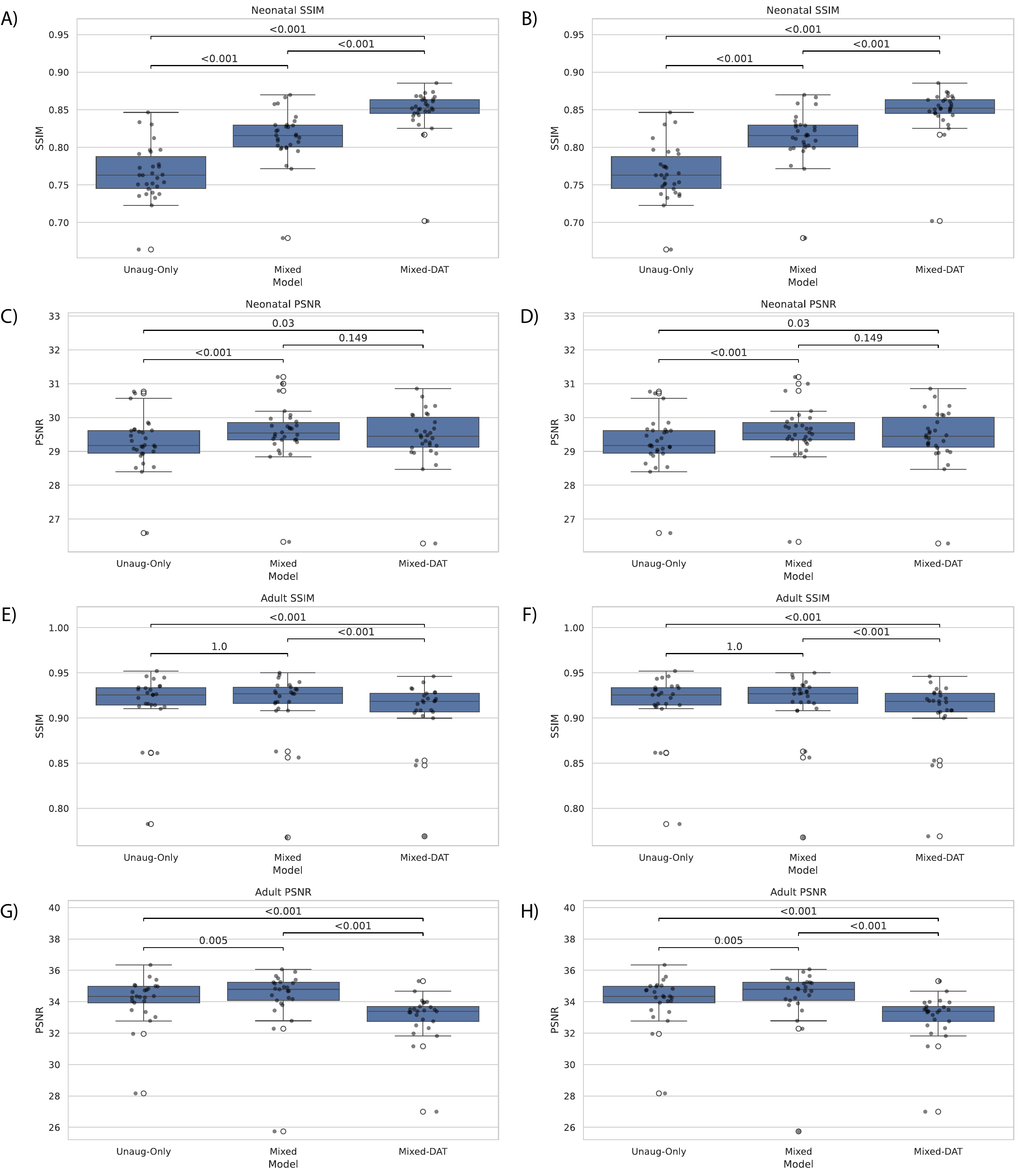}   
    \linespread{1}
    \caption{
        Boxplots of reconstruction performance on neonatal (A-D) and adult (E-H) test sets for each model under $R=4$ (A, C, E, and G) and $R=8$ (B, D, F, and H) Cartesian undersampling. Metrics are computed on a per-volume basis. Boxes represent the interquartile range (IQR), with the median indicated by the central line. Statistical significance was assessed using Bonferroni-corrected Wilcoxon signed-rank tests following a significant Friedman test. Brackets indicate statistical significance.
    }
    \label{fig:reconstruction-metrics-boxplots}
\end{figure}
%% RF: I really don't know why we convert good old p-values to a series of asterixs? My preference is alway to report the p-value and not the *-code. perhaps with bolding to indicate those that are significant.
%% RF: Labels on horizontal axis are not consistent with text and shoud also be defiend in the caption. You were calling these regimes earlier and NOT models?

Table \ref{tab:results_combined} summarizes reconstruction performance across all training regimes on neonatal and adult test sets. On the neonatal test set, both Mixed and Mixed-DAT outperformed Unaug-Only across both acceleration factors. At $R=4$, Mixed-DAT achieved the highest mean SSIM and PSNR (Figure \ref{fig:reconstruction-metrics-boxplots}, A and C). At $R=8$, Mixed-DAT achieved the highest mean SSIM, while Mixed achieved the highest PSNR (Figure \ref{fig:reconstruction-metrics-boxplots}, B and D). On the adult test set, the Unaug-Only and Mixed models performed similarly across both acceleration factors (Figure \ref{fig:reconstruction-metrics-boxplots}, E-H). Mixed-DAT performed marginally worse on adult data, especially when evaluated using PSNR (Figure \ref{fig:reconstruction-metrics-boxplots}, G-H).

\subsection{Effect of Domain-Adversarial Training}

To isolate the effect of DAT, we compared the performance of the Mixed and Mixed-DAT regimes, which differed only in the inclusion of the adversarial objective. We compared their performance on the neonatal test set as a measure of generalizability and on the adult test set as a measure of source domain retention. At $R=4$, when tested on neonatal data, the Mixed-DAT model produced reconstructions with significantly higher mean SSIM and PSNR than those produced by the Mixed model (Table \ref{tab:results_combined}). At $R=8$, the Mixed-DAT model produced reconstructions with significantly higher mean SSIM values but significantly lower PSNR values than those produced by the Mixed model. These trends are reflected in the distribution of performance across volumes (Figure \ref{fig:reconstruction-metrics-boxplots}, A-D), where Mixed-DAT reconstructions were consistently of higher quality. When evaluated on adult data, the Mixed-DAT model produced reconstructions with significantly lower mean SSIM and PSNR values than those produced by the Mixed model, demonstrating some trade-off in adult test set reconstruction performance. 

\subsection{Latent Feature Representation Analysis}

\begin{figure}
    \centering
    \includegraphics[width=\linewidth]{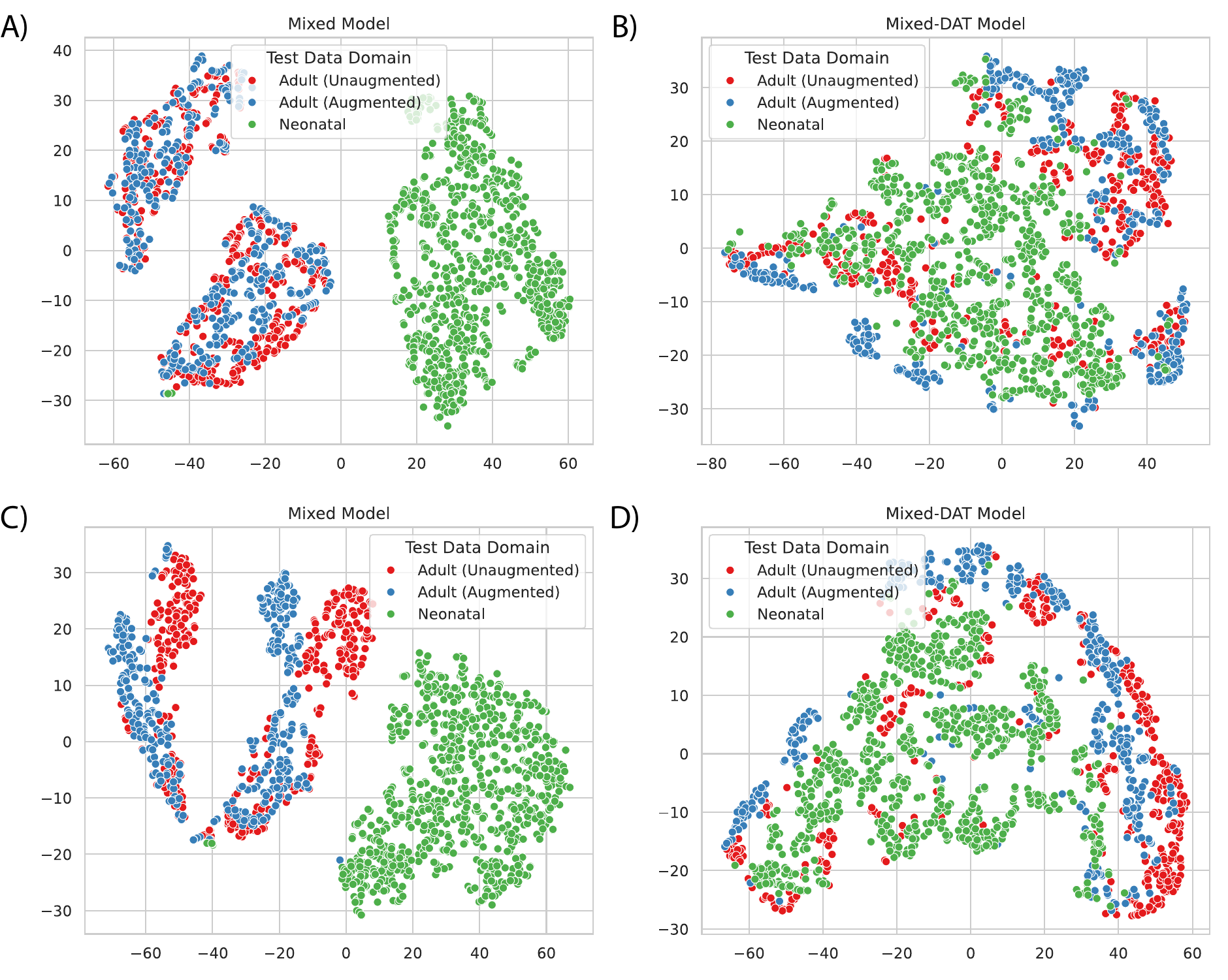}   
    \linespread{1}
    \caption{
        t-SNE visualization of latent feature representations extracted from the bottleneck layer of the final refinement module of the Mixed and Mixed-DAT models for unaugmented adult, augmented adult, and neonatal test samples. Panels A-B show $R=4$; panels C-D show $R=8$.
    }
    \label{fig:tsne-plots}
\end{figure}

To assess whether DAT altered the latent feature representations, we visualized bottleneck features from unaugmented adult, augmented adult, and neonatal test samples using t-SNE (Figure \ref{fig:tsne-plots}). The unaugmented and augmented adult data features extracted from the Mixed model showed partial separation, especially under $R=8$ undersampling (Figure \ref{fig:tsne-plots}, C), and the neonatal data features formed a cluster separate from both adult-derived training domains. The same unaugmented and augmented adult data features extracted from the Mixed-DAT model were slightly more overlapping, consistent with the intended effect of the adversarial objective. More notably, the neonatal data features extracted from the Mixed-DAT model overlapped substantially more with both adult-derived feature representations.

%% RF Are these M and M-DAT models or regimes?

\subsection{Qualitative Reconstruction Results}

\begin{figure}
    \centering
    \includegraphics[width=\linewidth]{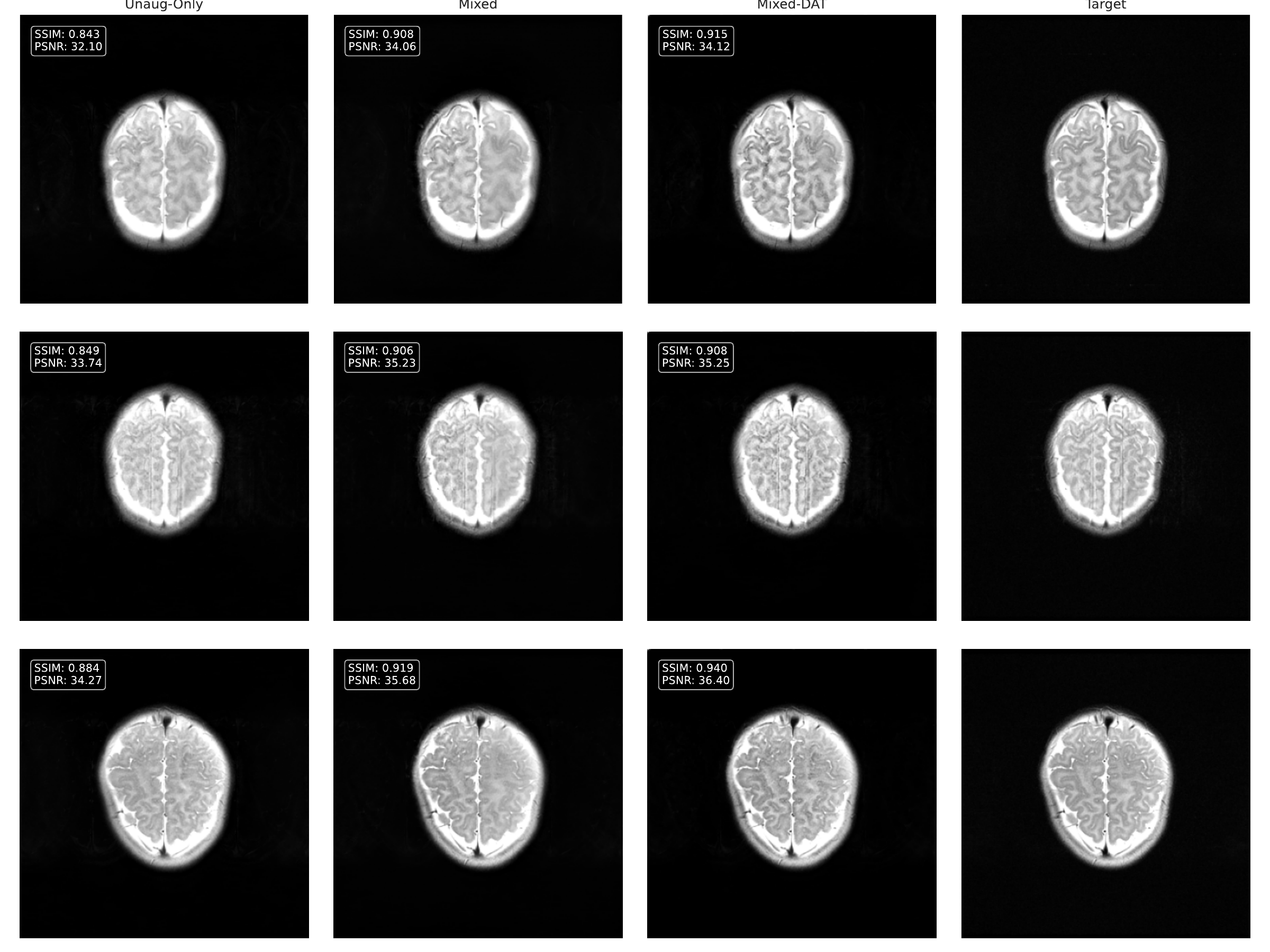} 
    \linespread{1}
    \caption{
        Representative neonatal reconstructions under $R=4$ retrospective Cartesian undersampling. Reconstructions generated from the Unaug-Only, Mixed, and Mixed-DAT models are displayed in the first three columns, respectively. Target reconstructions, generated from fully sampled data, are placed in the fourth column.
    }
    \label{fig:reconstruction-examples-neonatal-r4}
\end{figure}
%% RF: Labels the rows as a differnet patient or say this in the caption. What is a target - you say "fully sampled" in the caption. Make sure (here and other figures) that you nice plots align with the caption and the text - i find sometimes your labels don't match.

\begin{figure}
    \centering
    \includegraphics[width=\linewidth]{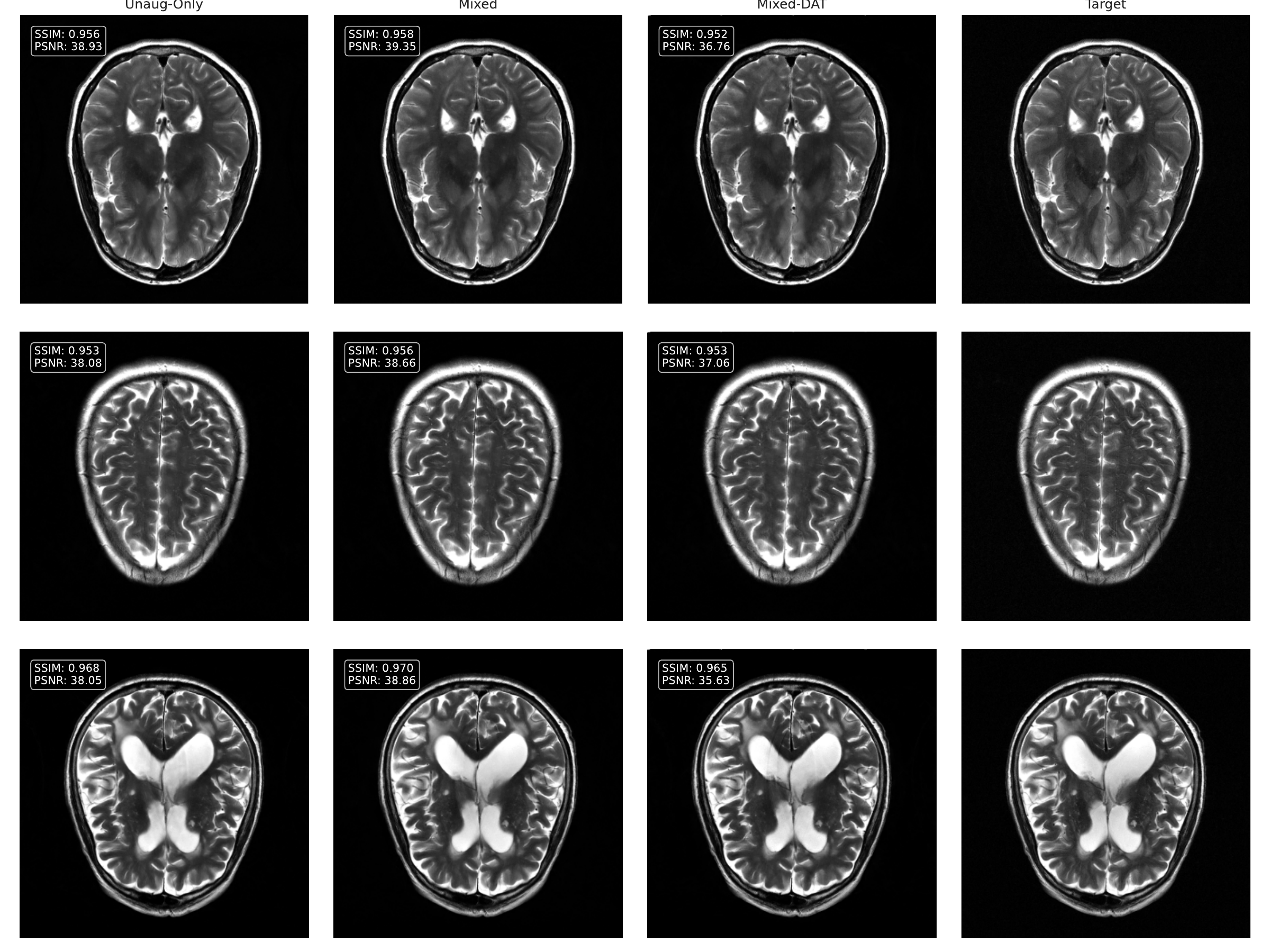}   
    \linespread{1}
    \caption{
        Representative adult reconstructions under $R=4$ retrospective Cartesian undersampling. Reconstructions generated from the Unaug-Only, Mixed, and Mixed-DAT models are displayed in the first three columns, respectively. Target reconstructions, generated from fully sampled data, are placed in the fourth column.
    }
    \label{fig:reconstruction-examples-adult-r4}
\end{figure}

\begin{figure}
    \centering
    \includegraphics[width=\linewidth]{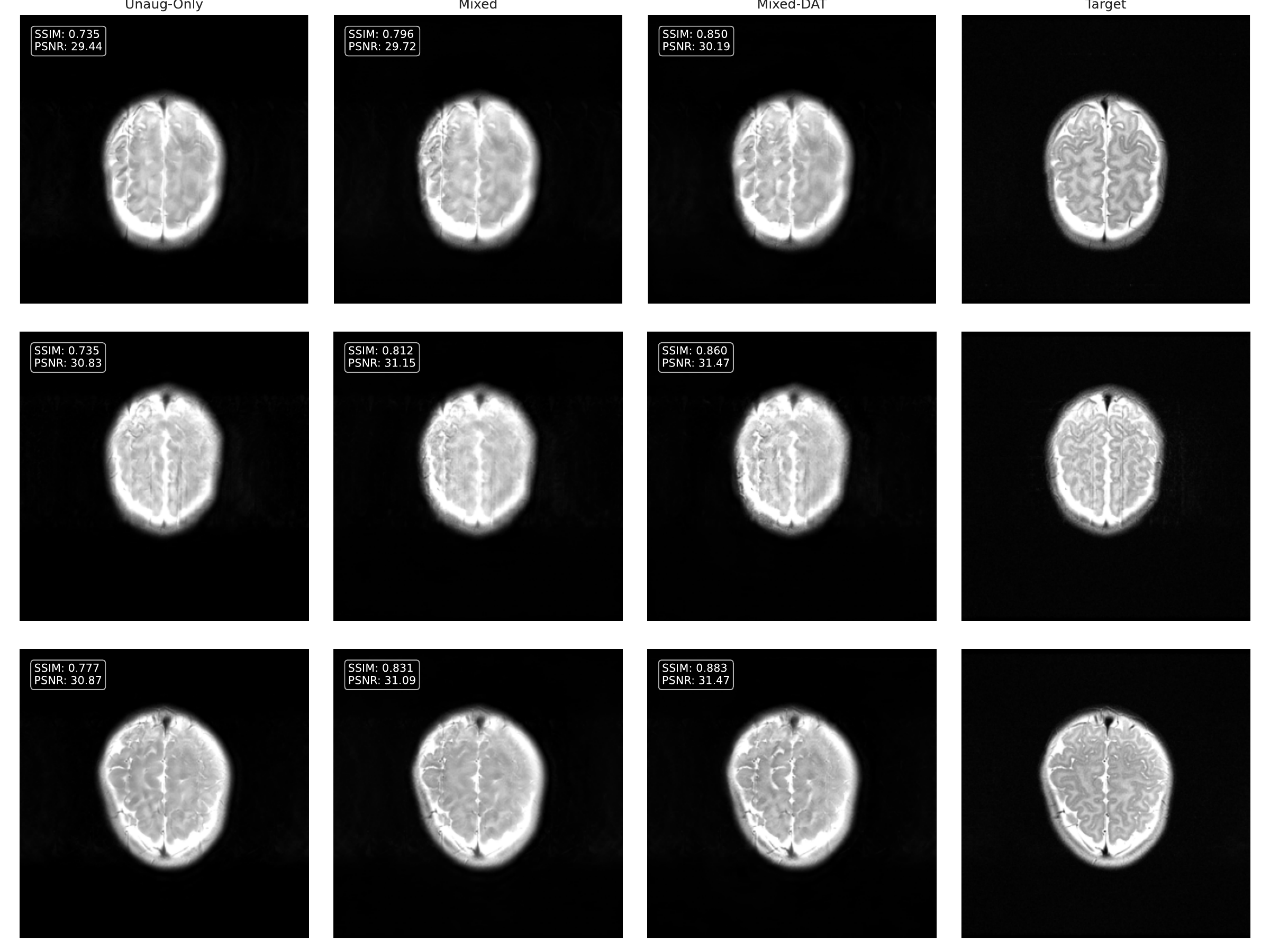} 
    \linespread{1}
    \caption{
        Representative neonatal reconstructions under $R=8$ retrospective Cartesian undersampling. Reconstructions generated from the Unaug-Only, Mixed, and Mixed-DAT models are displayed in the first three columns, respectively. Target reconstructions, generated from fully sampled data, are placed in the fourth column.
    }
    \label{fig:reconstruction-examples-neonatal-r8}
\end{figure}

\begin{figure}
    \centering
    \includegraphics[width=\linewidth]{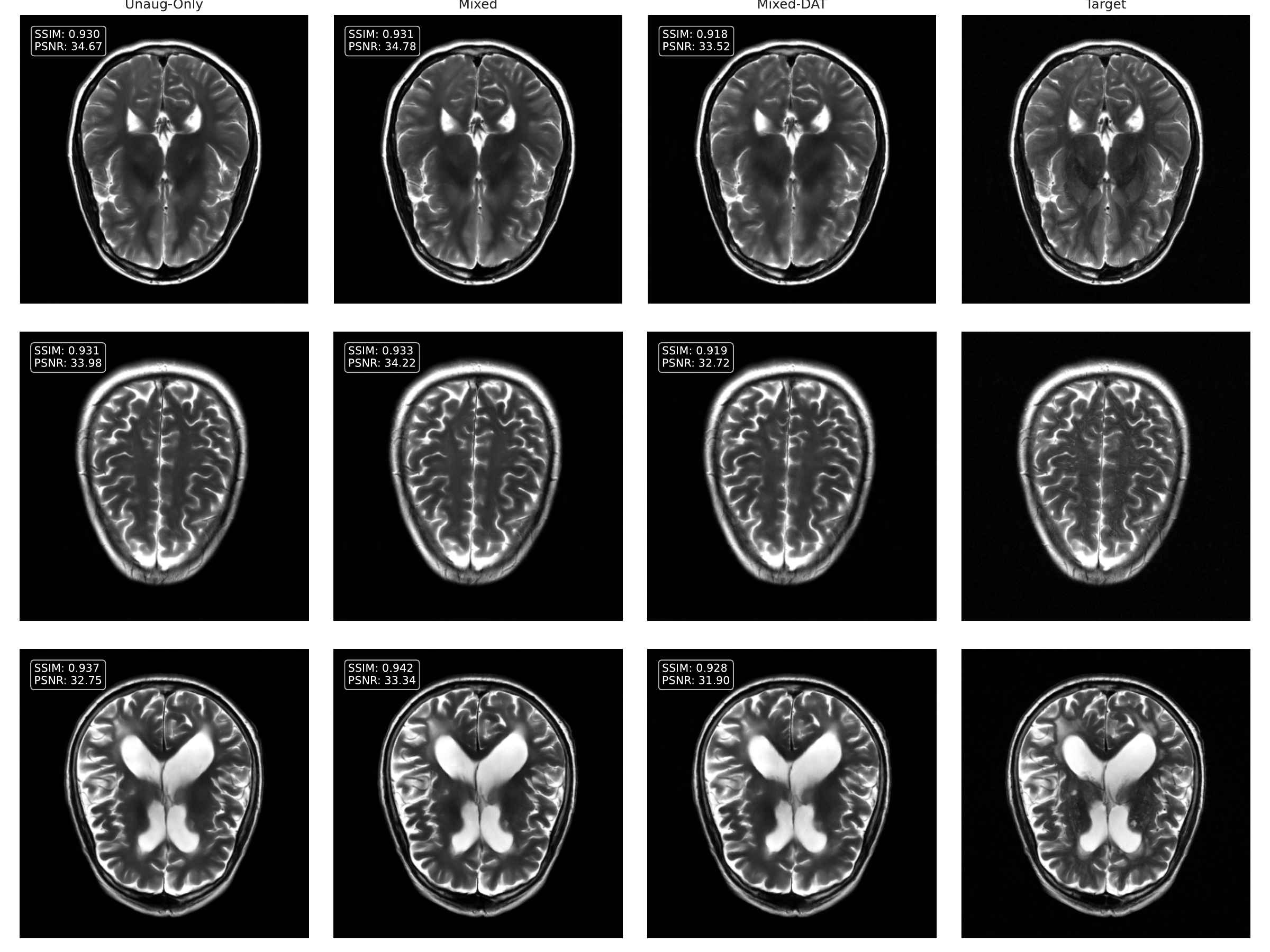}   
    \linespread{1}
    \caption{
        Representative adult reconstructions under $R=8$ retrospective Cartesian undersampling. Reconstructions generated from the Unaug-Only, Mixed, and Mixed-DAT models are displayed in the first three columns, respectively. Target reconstructions, generated from fully sampled data, are placed in the fourth column.
    }
    \label{fig:reconstruction-examples-adult-r8}
\end{figure}

Representative reconstructions of samples from the neonatal test set are shown in Figures \ref{fig:reconstruction-examples-neonatal-r4}-\ref{fig:reconstruction-examples-adult-r8}. The qualitative assessment by expert LL was largely consistent with the quantitative results but did not fully mirror them at $R=4$. According to LL, all models adequately reconstructed $R=4$ undersampled neonatal data for clinical interpretation (Figure \ref{fig:reconstruction-examples-neonatal-r4}). The Mixed-DAT model was rated as producing the highest quality reconstructions, followed by the Unaug-Only and Mixed models (Figure \ref{fig:reconstruction-examples-neonatal-r4}). Also, the Unaug-Only reconstructions had higher contrast and sharper contrast differentiation than the Mixed model (Figure \ref{fig:reconstruction-examples-neonatal-r4}). LL determined that no models adequately reconstructed $R=8$ undersampled neonatal data for clinical interpretation. However, Mixed-DAT reconstructions were observed to have less visually apparent blurring and marginally improved white and grey matter contrast differentiation (Figure \ref{fig:reconstruction-examples-neonatal-r8}). The qualitative differences between the adult data reconstructions produced by the Unaug-Only, Mixed, and Mixed-DAT models were less apparent. Generally, all models produced high-quality reconstructions of adult data, although modest decreases in sharpness were observed in the reconstructions of $R=8$ undersampled data (Figure \ref{fig:reconstruction-examples-adult-r8}).

% ======================================================================
\section{Discussion}
% ======================================================================

In this study, we investigated the impacts of contrast-informed data augmentation and DAT on the generalizability of DL-based MR reconstruction models trained on adult data when applied to neonatal data. The use of DAT was motivated by our hypothesis that, because the applied augmentations were designed to imitate neonatal data characteristics, models trained to emphasize augmentation-invariant feature representations would generalize better to real neonatal data. We showed empirically that incorporating data augmented to mimic the domain shift from the source (\textit{i.e.}, adult) to the target (\textit{i.e.}, neonatal) domain improved the model's generalizability to real neonatal data compared to adult-only training. Adding DAT to the mixed training regime further improved reconstruction of test neonatal data, with the clearest benefit observed under $R=8$ undersampling. 

The first principal contribution of this study is the use of augmentations specifically designed to minimize the domain shift between training and testing data. The use of contrast-modulating augmentations to address the adult-neonatal domain shift in MR imaging is not novel, with previous work demonstrating its potential in skull stripping \cite{omidi2025improving,omidi2024unsupervised}. However, to our knowledge, incorporating neonatal-imitating contrast augmentations into training data to improve the generalizability of MR reconstruction models to real neonatal data has not previously been investigated. The observed improvement in reconstruction performance from models trained exclusively on unaugmented adult data to those trained with an equal proportion of unaugmented and augmented adult data suggests that neonatal-imitating augmentations productively expand the training data distribution such that it improves generalization to neonatal data. However, while contrast-informed, our augmentation protocol is rudimentary and does not fully replicate neonatal MR imaging data. As such, further investigation is necessary before attributing the observed performance gains specifically to the neonatal likeness of the proposed augmentations. Such investigations might include comparisons to more general augmentations not informed by neonatal MR image characteristics, and to more refined augmentation protocols that better simulate neonatal MR image characteristics.

The second principal contribution of this study is the addition of DAT to a DL-based MR reconstruction model. Previous work has proposed adversarial feature alignment for cross-site MR reconstruction in a federated learning framework \cite{guo_multi-institutional_2021}, establishing adversarial domain alignment as a relevant strategy for reconstruction under domain shift. However, using DAT to improve generalizability from adult to neonatal data has not, to our knowledge, been explored. In our study, adding DAT to the mixed training regimen improved reconstruction of neonatal test data compared to mixed training alone. At $R=4$, Mixed-DAT improved both SSIM and PSNR, and at $R=8$, Mixed-DAT improved SSIM but slightly worsened PSNR. Because the discriminator was trained to classify data as unaugmented or augmented, these findings suggest that training the model to reduce sensitivity to augmentation-defined domain differences can improve generalization to real neonatal data. Moreover, the representation analyses indicated that DAT altered the models' learned feature representations. While overlap between unaugmented and augmented adult features increased modestly from Mixed to Mixed-DAT, the more notable difference was observed for real neonatal samples: the neonatal features extracted from the Mixed model formed a cluster distinct from both adult-derived training domains, whereas the neonatal features extracted from the Mixed-DAT model overlapped substantially with unaugmented and augmented adult features. Although qualitative and limited by the interpretability of t-SNE embeddings, this finding is consistent with DAT reducing the feature-space separation between real neonatal data and the adult-derived training distributions. Together, the reconstruction and representation analyses are consistent with DAT improving generalization by reducing model sensitivity to augmentation-defined domain differences. However, these results do not establish that the observed neonatal reconstruction gains were caused specifically by the learning of domain-invariant features; such a mechanistic interpretation would require further ablation and correlation analyses. Nevertheless, because the discriminator was trained on unaugmented and augmented adult data, the increased overlap of their features with neonatal features suggests that the augmentation-defined adversarial objective captured aspects of the real adult-to-neonatal shift that were relevant to the learned reconstruction representation.

Notably, SSIM and PSNR did not follow the same trends. Although adding DAT to the mixed training regime improved neonatal SSIM at both $R=4$ and $R=8$, it did not always improve PSNR: DAT improved PSNR at $R=4$ but not at $R=8$. The divergent trends in SSIM and PSNR suggest that DAT shifted reconstruction to emphasize structural similarity. This trend may arise from SSIM's use as a loss function, rather than pixel-wise alternatives more closely related to PSNR. Future studies should investigate whether DAT preferentially improves metrics aligned with the reconstruction objective, at the expense of complementary quantitative measures not directly optimized during training. Furthermore, while this study determines reconstruction quality primarily by quantitative metrics, these metrics do not necessarily correlate with or establish clinical utility. For example, while Mixed-DAT produced higher-quality reconstructions of $R=8$ undersampled neonatal data, they remained of insufficient quality for confident clinical interpretation. As such, while further modification of loss functions or general refinement of training regimes may produce models with more convergent quantitative metric trends (\textit{e.g.}, for SSIM and PSNR), the most consequential measure of performance is ultimately its utility for clinical interpretation.   

Related to divergent SSIM and PSNR trends, we observed a slight reduction in adult-domain reconstruction performance under DAT compared to mixed training, most notably in PSNR. This finding suggests a possible trade-off in which DAT improves neonatal generalization while modestly reducing adult-domain reconstruction performance. This source-domain trade-off may arise from DAT suppressing adult-domain-specific information that is useful for the reconstruction task. However, evaluating this hypothesis would require further study. This slight observed trade-off in adult reconstruction performance, especially as measured by PSNR, may be acceptable because our primary objective was to improve reconstruction in the data-scarce target domain--neonatal MR--rather than to jointly optimize a single model for both adult and neonatal reconstruction performance. Nevertheless, this finding underscores the impact of the strength and scheduling of the adversarial objective. We did not exhaustively evaluate these and other relevant tunable model and training parameters due to high computational cost, but given our findings, future work should systematically investigate this balance. Such analyses might, for example, include varying the adversarial weighting parameter, domain loss weight, and domain discriminator architecture to identify training regimes that maximize both DAT-associated generalization to neonatal data and performance retention on adult data.

This study also has several more general limitations. First, a relatively small, single-site dataset represented the neonatal domain, and the generalizability of the findings to other neonatal cohorts, especially with respect to specific developmental stages, remains unestablished. Second, the contrast-informed augmentation protocol was--to maximize computational efficiency--intentionally simple and approximated only select characteristics of the adult-to-neonatal domain shift. For example, it did not explicitly model anatomical differences or motion artifacts. Third, all models were trained and evaluated on retrospectively undersampled data. Assessing the potential clinical relevance of the proposed methods will require prospectively accelerated acquisitions. Finally, while the reconstruction and representation analyses demonstrate the potential utility of DAT in adult-to-neonatal generalization, they do not fully explain the mechanism through which DAT improves neonatal reconstruction performance. Future work should therefore combine more extensive feature-space analyses and targeted ablation experiments to clarify the mechanism of DAT-associated generalization, alongside prospective evaluation to determine whether these gains translate to clinically meaningful improvements in neonatal MR imaging.

% ======================================================================
\section{Conclusions}
% ======================================================================

Contrast-informed augmentation improved the generalizability of adult-trained DL-based MR reconstruction models to neonatal MR data. Adding a domain-adversarial training objective to the mixed training regime further improved neonatal reconstruction performance, particularly for SSIM under $R=8$ undersampling, although modest trade-offs in adult-domain performance were observed. Analysis of the latent feature representations suggested reduced feature-space separation between real neonatal data and adult-derived training distributions, consistent with the intended effect of the adversarial objective. These findings support contrast-informed augmentation and domain-adversarial training as promising strategies for improving reconstruction robustness in underrepresented MR imaging domains, and highlight the need for prospective validation and further investigation of the mechanisms underlying DAT-associated generalization in DL-based MR reconstruction.

% ======================================================================
\section{Acknowledgments}
% ======================================================================

SM is supported by a Natural Sciences and Engineering Research Council (NSERC) Canada Graduate Scholarship Master's (CGS-M) and an Alberta Innovates Graduate Student Scholarship. Dr. Souza is supported by a NSERC Discovery Grant  and the Alberta Innovates LevMax Health Programs. Dr. Frayne is supported by a second NSERC Discovery Grant. Our team is also supported by the Digital Research Alliance of Canada compute infrastructure, and funding from the Alberta Children’s Hospital Foundation, Calgary Health Foundation and the Canada First Research Excellence Fund (One Child Every Child). 

% ======================================================================
% \section{References}
% ======================================================================

%% RF: Some of the references are incomplete.
\bibliography{references.bib} 

% ======================================================================
\section{Tables and Captions}
% ======================================================================

% ======================================================================

% ======================================================================
\section{Figure Captions}
% ======================================================================

% ======================================================================

\end{document}